%% file: PaperForReview.tex
\documentclass[10pt,twocolumn,letterpaper]{article}

\usepackage{cvpr}      

\usepackage{graphicx}
\usepackage{amsmath}
\usepackage{amssymb}
\usepackage{booktabs}
\input{commands}

\usepackage[pagebackref,breaklinks,colorlinks]{hyperref}

\usepackage[capitalize]{cleveref}
\crefname{section}{Sec.}{Secs.}
\Crefname{section}{Section}{Sections}
\Crefname{table}{Table}{Tables}
\crefname{table}{Tab.}{Tabs.}

\def\cvprPaperID{23} 
\def\confName{CVPR}
\def\confYear{2022}

\begin{document}

\title{On Fragile Features and Batch Normalization in Adversarial Training}

\author{Nils Philipp Walter \qquad David Stutz \qquad Bernt Schiele\\
Max Planck Institute for Informatics, Saarland Informatics Campus, Saarbr\"{u}cken\\
{\tt\small \{nwalter,david.stutz,schiele\}@mpi-inf.mpg.de}	
}
\maketitle

\begin{abstract}
    Modern deep learning architecture utilize \emph{batch normalization (BN)} to stabilize training and improve accuracy.
    It has been shown that the BN layers alone are surprisingly expressive.
    In the context of robustness against adversarial examples, however, BN is argued to increase vulnerability.
    That is, BN helps to learn \emph{fragile} features.
    Nevertheless, BN is still used in adversarial training, which is the de-facto standard to learn \emph{robust} features.
    In order to shed light on the role of BN in adversarial training, we investigate to what extent the expressiveness of BN can be used to ``robustify'' fragile features in comparison to random features.
    On CIFAR10, we find that adversarially fine-tuning just the BN layers can result in non-trivial adversarial robustness. 
    Adversarially training \emph{only} the BN layers from scratch, in contrast, is not able to convey meaningful adversarial robustness.
    Our results indicate that fragile features can be used to learn models with moderate adversarial robustness, while random features cannot.
\end{abstract}
\vspace{-0.5cm}
\section{Introduction}
Deep Neural Networks (DNNs) have set the state-of-the-art for many tasks in computer vision.
Almost all DNN architectures make use of \emph{batch normalization (BN)} \cite{Ioffe2015BatchNA} to stabilize the training procedure.
In particular, BN plays an important role in being able to train particularly deep architectures such as ResNets \cite{HeCVPR2016}.
Besides that, BN layers have been shown to be surprisingly expressive compared 
to convolution and fully connected layers:
\cite{frankle2021training} showed that only training the BN layers is sufficient to achieve non-trivial accuracy for very deep ResNets.
The fact that BN-layers are equipped with significantly fewer parameters makes the results particularly surprising.

On the other hand, BN has also been argued to increase vulnerability against adversarial examples \cite{GallowayARXIV2019}, imperceptibly perturbed images causing mis-classification \cite{szegedy2014intriguing}.
The work in \cite{XieARXIV2019} suggests that the statistics of the BN layers could be responsible not only for adversarial vulnerability but also for the reduced (clean) accuracy frequently observed, due to the fact that the
 activation statistics of clean examples and adversarial examples strongly differ \cite{rev1, rev2}.
Beyond BN, it is argued \cite{ilyas2019adversarial} that adversarial examples are a consequence of so called \emph{fragile} features.
A fragile feature is believed to pick up spurious correlations, that are exploited by the attacker to craft adversarial examples.
The most common way to tackle adversarial vulnerability is adversarial training,
which learns \emph{robust} features and has been shown to ignore spurious correlations.
However, \cite{sehwag2020hydra} also shows that it is possible to extract robust networks from normally trained ones.
This indicates that there are robust features ``hidden'' among fragile ones.

Complementary to these works, we investigate how adversarial \emph{fine-tuning} of the BN layers can be used to improve adversarial robustness.
For that we adversarially tune the BN parameters (and statistics) starting with a normally trained, non-robust base model.
On CIFAR10 \cite{Krizhevsky2009}, fine-tuning statistics \emph{and} parameters leads to non-trivial adversarial robustness at the cost of reduced clean accuracy.
These experiments indicate that training just the BN parameters allows to utilize fragile features for adversarial robustness -- at least to some extent.
As no robustness can be achieved when training \emph{only} the BN layers (from scratch), random feature are shown to be useless for robustness.
With these experiments, our work supports existing evidence that there exist robust features among fragile ones \cite{sehwag2020hydra}.
However, adversarial training seems to learn \emph{additional} robust features not learned using standard training.

\section{Related Work}

\textbf{Adversarial Training:}
Following \cite{MadryICLR2018}, adversarial examples can be obtained by maximizing cross-entropy loss, \ie, $\max_{\delta \in \mathcal{S}} L(f(x+\delta), y)$ where $L$ computes the loss between the classifier's output $f(x + \delta)$ and the ground truth label $y$.
Then, adversarial training can be formulated as a min-max learning problem $\min_{\theta} \mathbb{E}_{(x, y) \sim \mathcal{D}}\left[\max_{\delta \in \mathcal{S}} L(f_\theta(x+\delta), y)\right]$.
For evaluation, ensembles of different attacks has become standard \cite{croce2020reliable}.

\textbf{Fragile and Robust Features:} \cite{tsipras2019robustness} suggests that recent
classifiers have to rely on so called \textit{fragile} features to obtain high accuracy.
However, these spurious correlations can be exploited by adversarial examples.
Thus, improving adversarial robustness naturally leads to reduced (clean) accuracy \cite{ilyas2019adversarial,StutzCVPR2019}.
While \cite{sehwag2020hydra} shows that robust sub-networks exist within normal networks, it remains largely unclear to \emph{what extent} robust features replace or built-upon fragile features to improve adversarial robustness.

\textbf{Batch Normalization (BN):}
It is still poorly understood how exactly BN helps training \cite{SanturkarNIPS2018}.
Nevertheless, BN is known to be very expressive itself \cite{frankle2021training}:
training \emph{only} the BN parameters (and updating the statistics) is sufficient to obtain non-trivial accuracy.
Moreover, units/channels with small BN parameters can be removed without affecting accuracy,
indicating that BN learns to ``turn off'' specific (random) features.
In the context of adversarial robustness, BN is argued to increase vulnerability \cite{GallowayARXIV2019}
and the statistics are known to be very different when training on adversarial examples \cite{XieARXIV2019}.
We study how fragile features are utilized in adversarial fine-tuning in comparison to random features.


\section{Fragile Features and Batch Normalization in Adversarial Training}

We aim to understand to what extent non-trivial robustness is possible based on fragile features compared to random features.
To this end, we adversarially fine-tune normally trained models and evaluate robustness compared to adversarially training the BN (and only the BN) layers from scratch.
We briefly include the high-level idea of our experiments in Section \ref{subsec:methdology},
before discussing experimental setup in Section \ref{subsec:setup} and presenting our results in Section \ref{subsec:results}

\subsection{Methodology}
\label{subsec:methdology}

\begin{table}[t]
\caption{Overview of training configurations: We consider fine-tuning only the BN statistics (\BNStats), only the BN parameters (\BNOnlyParams), or the statistics \emph{and} parameters (\BNParams). Additionally, we consider training logits and the first convolutional layer. We also consider normally/adversarially training \emph{only} the BN layers from scratch (\BNOnly).}
\label{table:configurations}
\vspace*{-8px}
\centering
\small
\begin{tabular}{@{\hskip 6px}c@{\hskip 6px}l@{\hskip 6px}c@{\hskip 6px}c@{\hskip 6px}c@{\hskip 6px}r@{\hskip 6px}}
\hline
& Model & all & \begin{tabular}{@{}c@{}}BN\\params\end{tabular} & \begin{tabular}{@{}c@{}}BN\\stats\end{tabular} & \%params \\ 
\hline
\multirow{2}{*}{From scratch} & \Normal/ \Adv & \cmark &&& 100.00\\
& +\BNOnly & \xmark & \xmark & \cmark & 0.15\\
\hline
\multirow{2}{*}{Fine-tune} & \BNStats & \xmark & \xmark & \cmark & (0.15)\\ 
& \BNParams & \xmark & \cmark & \cmark & 0.15 \\
\hline
\end{tabular}
\vspace*{-0.15cm}
\end{table}

This study is mostly based on two observations: First, \cite{frankle2021training} showed that training only
the BN parameters (and statistics) is sufficient to achieve decent accuracy.
Furthermore they argued that BN learns to identify and ``turn off'' useless random features.
Second, it is argued that adversarial vulnerability is caused by \emph{fragile features} -- spurious correlations that are useful
for accuracy but can be exploited by an adversary.
Note that fragile features are not restricted to the image domain as suggested in \cite{tsipras2019robustness}
because they propagate into deeper layers when used.

In order to quantify to what degree these fragile features prevent adversarial robustness, we conduct two main experiments:
We \emph{fine-tune} a normally trained model using adversarial training \cite{MadryICLR2018}
by only updating the BN parameters (and statistics).
This means all parameters except the BN parameters are frozen.
The motivation is that adversarial training will try to avoid or deactivate non-robust, \ie, fragile, features in order to improve adversarial robustness.
We emphasize that for, \eg, a ResNet-20, the BN parameters only make up 0.15\% of the parameters.
Then, we also consider training only the BN parameters (and statistics) \emph{from scratch} using adversarial training.
For standard accuracy, random features are shown to be helpful.
Thus, this acts as a baseline to check whether adversarial training of only the BN parameters on top of fragile features works better than on random ones.



\subsection{Experimental Setup}
\label{subsec:setup}

\begin{table}[t]
\caption{Main quantitative results for adversarial fine-tuning of BN layers: following the naming of Table \ref{table:configurations},
we report clean error (in \%) and robust error (against AutoAttack, in \%) on test examples for $L_\infty$ attacks with $\epsilon = \nicefrac{8}{255}$.
The baseline \Adv model improves robust error significantly, at the cost of increased clean error.
While \BNStats and \Adv \BNOnly are unable to obtain non-trivial robustness, \BNParams reduces robust error significantly, even if not matching \Adv.
This indicates that robustness with fragile features is possible, but limited.} 
\label{table:main_summary}
\vspace*{-8px}
\centering
\small
\begin{tabular}{@{\hskip 6px}l@{\hskip 6px}c@{\hskip 6px}c@{\hskip 6px}}
\hline
Model & Test error &  Robust test error\\ 
\hline
\Normal & 4.11 & 99.8 \\ 
+\BNOnly & 46.11 &  99.7 \\ 
+\BNOnly + log + conv1 & 35.89 &  99.6 \\
\hline
\Adv & 17.67 & 58.4 \\
+\BNOnly & 43.33 &  98.0 \\ 
+\BNOnly + log + conv1 & 35.22 &  98.1 \\ 
\midrule
\BNStats (no reset) & 6.44 & 99.9 \\ 
\BNStats (reset) & 6.33 &  99.6 \\ 
\hline
\BNOnlyParams & 77.11 & 85.3 \\
\hline
\BNParams & 29.67 &  70.5 \\ 
\BNParams + log + conv1 & 26.89 &  68.2 \\
\hline
\end{tabular}
\vspace*{-0.15cm}
\end{table}

For simplicity, we use ResNet-20 \cite{HeCVPR2016} as a base model.
Similar to \cite{frankle2021training}, we place the BN layers before the activation layer, which leads to better performance according to \cite{preact}.
All our experiments are conducted on the CIFAR-10 dataset \cite{Krizhevsky2009}.
For the (normally trained) baseline model, we also employ AutoAugment \cite{cubuk2019autoaugment} to improve accuracy.

For adversarial fine-tuning, we use $10$ iterations of PGD with a $L_\infty$ perturbation budget of $\epsilon = \nicefrac{8}{255}$.
In contrast to the baseline model, we do not use AutoAugment, but rather follow related work and employ random crops and flips only \cite{MadryICLR2018}.
We note that accurate classifiers are argued to rely on spurious correlations, justifying our use of AutoAugment for the baseline model.
For adversarial training, however, AutoAugment is not reported to improve robustness.
At test time, we use Autoattack \cite{croce2020reliable} for robustness evaluation on the first $1000$ test examples.

We report results for several training configurations, see Table \ref{table:configurations}:
Given a normally trained baseline model (\Normal), we adversarially fine-tune
it by only updating the BN statistics (\BNStats) or both the statistics and parameters (\BNParams).
In the latter case, we might also learn the logit or first convolutional layer.
As baseline, we consider a adversarially trained model (\Adv).
Finally, we follow \cite{frankle2021training} and consider adversarially training only the BN layers (from scratch, \Adv + \BNOnly).

\subsection{Results}
\label{subsec:results}

\begin{figure}[t]
    \vspace*{-0.2cm}
    \centering
    \includegraphics[scale=.3]{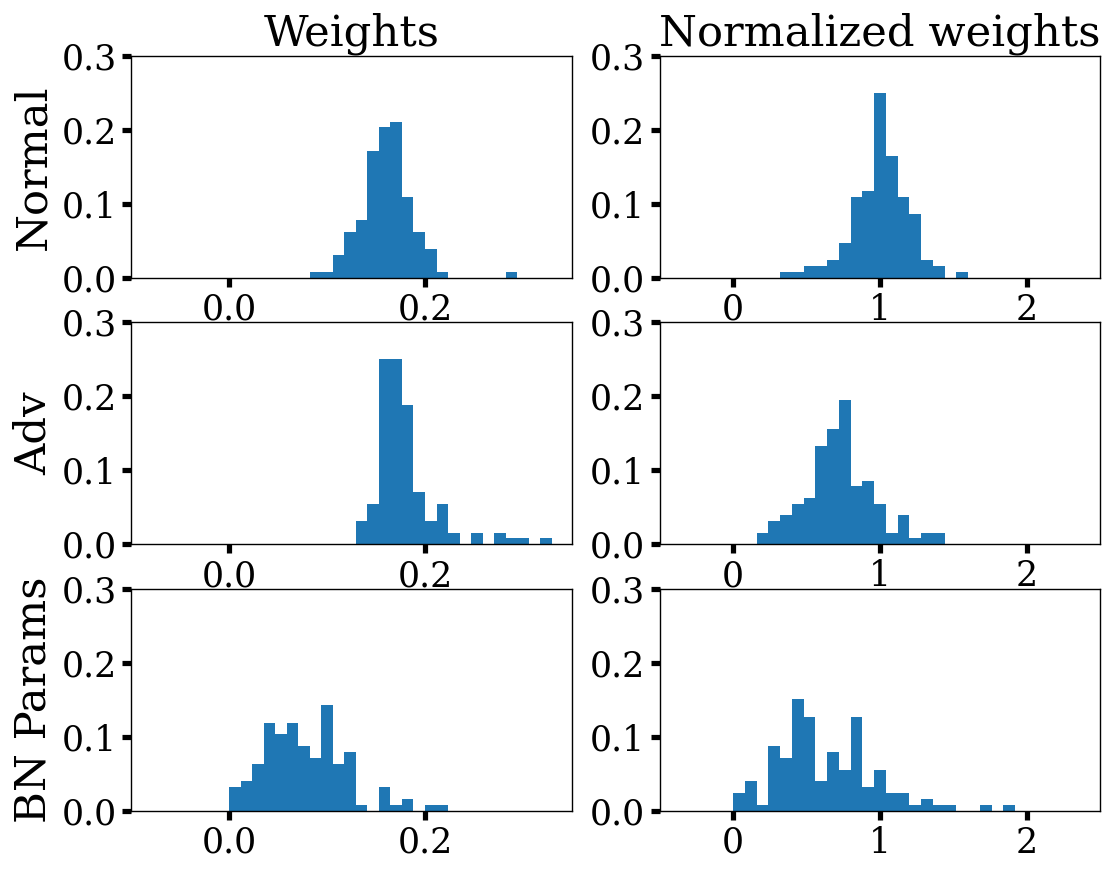}
    \vspace*{-9px}
    \caption{We plot histograms of the BN parameters (8th layer) and their normalized versions according to Equation \ref{eq:normalized} for \Normal, \Adv and \BNParams. On average, 
    \BNParams reduces the normalized weights $m$ compared to the baseline models.
    This indicates that robustness is achieved by ``turning off'' a non-significant portion of the fragile features.}
    \label{fig:parameters}
    \vspace*{-0.5cm}
\end{figure}

We report our main results in Table \ref{table:main_summary}.
We first consider only fitting the BN statistics \emph{or} parameters, before considering to update both, potentially with the logit and first convolutional layer.
This is then compared to adversarially training BN (and only BN) from scratch.
This allows us to draw conclusions about the usefulness of fragile and random features for adversarial robustness.
As shown in Table \ref{table:main_summary}, our normally trained baseline (\Normal) achieves a clean error of $4.11\%$, but is not robust against adversarial examples (robust error $99.8\%$).
The adversarially trained baseline (\Adv) achieves $17.67\%$ clean error and improves robustness with $58.4\%$ robust error.
This reflects the commonly observed robustness-accuracy trade-off.

\textbf{Importance of BN Statistics:}
As a first experiment we (adversarially) fine-tune \emph{only} the BN-statistics (\ie, $\mu$ and $\sigma$). 
Note that there are as many statistics as BN parameters. 
The statistics are updated by simply forwarding adversarial examples.
While clean test error increases slightly, robust test error does \emph{not} improve.
We confirmed this for smaller values of $\epsilon$, as well.
Also, the cross-entropy loss on adversarial examples does not change significantly (measured using PGD as AutoAttack does \emph{not} maximize loss).
\emph{Not} updating the BN statistics and just learning the BN parameters, in contrast, improves adversarial robustness slightly, from $99.8$ to $85.3\%$ robust test error.
However, clean test error is nearly as high.
These results show that neither statistics nor parameters alone are capable of obtaining meaningful, non-trivial adversarial robustness.
We also did not find any recognizable difference in the statistics or parameter distributions across layers.
Moreover, learning the logit layer, to account for the updated statistics, did not help and results do not change when resetting the statistics before forwarding adversarial examples.

\begin{figure}[t]
    \vspace*{-0.1cm}
    \includegraphics[scale=0.19]{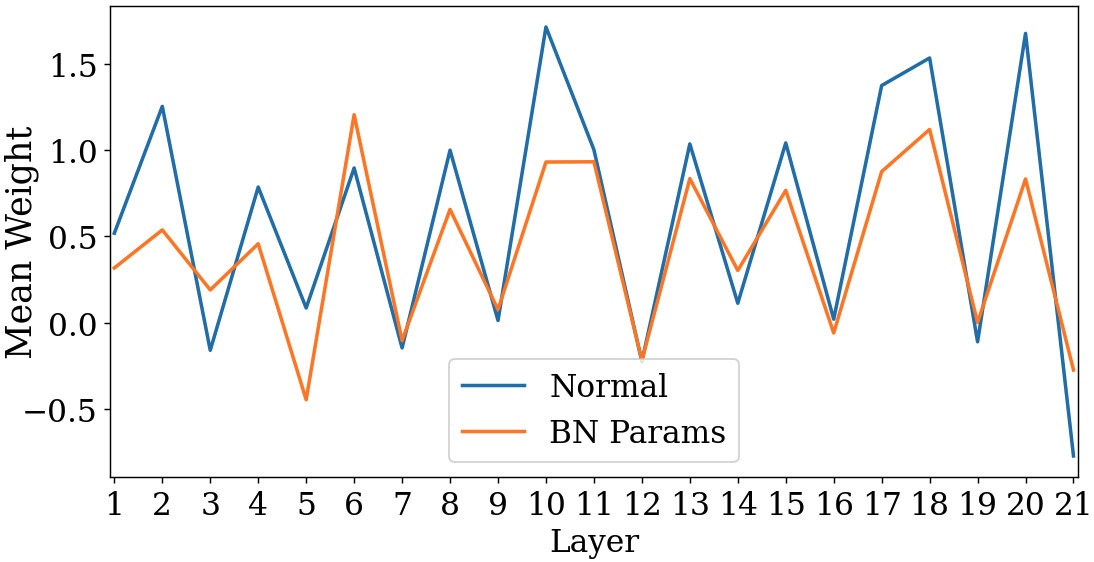}
    \vspace*{-10px}
    \caption{We plot the mean of the normalized weights per layer. On average, the yellow line is (\BNParams) below the blue
    line (\Normal), which indicates that adversarially fine-tuning the BN-layers tilts the normalized weights to be more 
    negative.\label{fig:meanweights}}
    \vspace*{-0.15cm}
\end{figure}

\begin{figure}[t]
    \vspace*{-0.1cm}
    \includegraphics[scale=0.23]{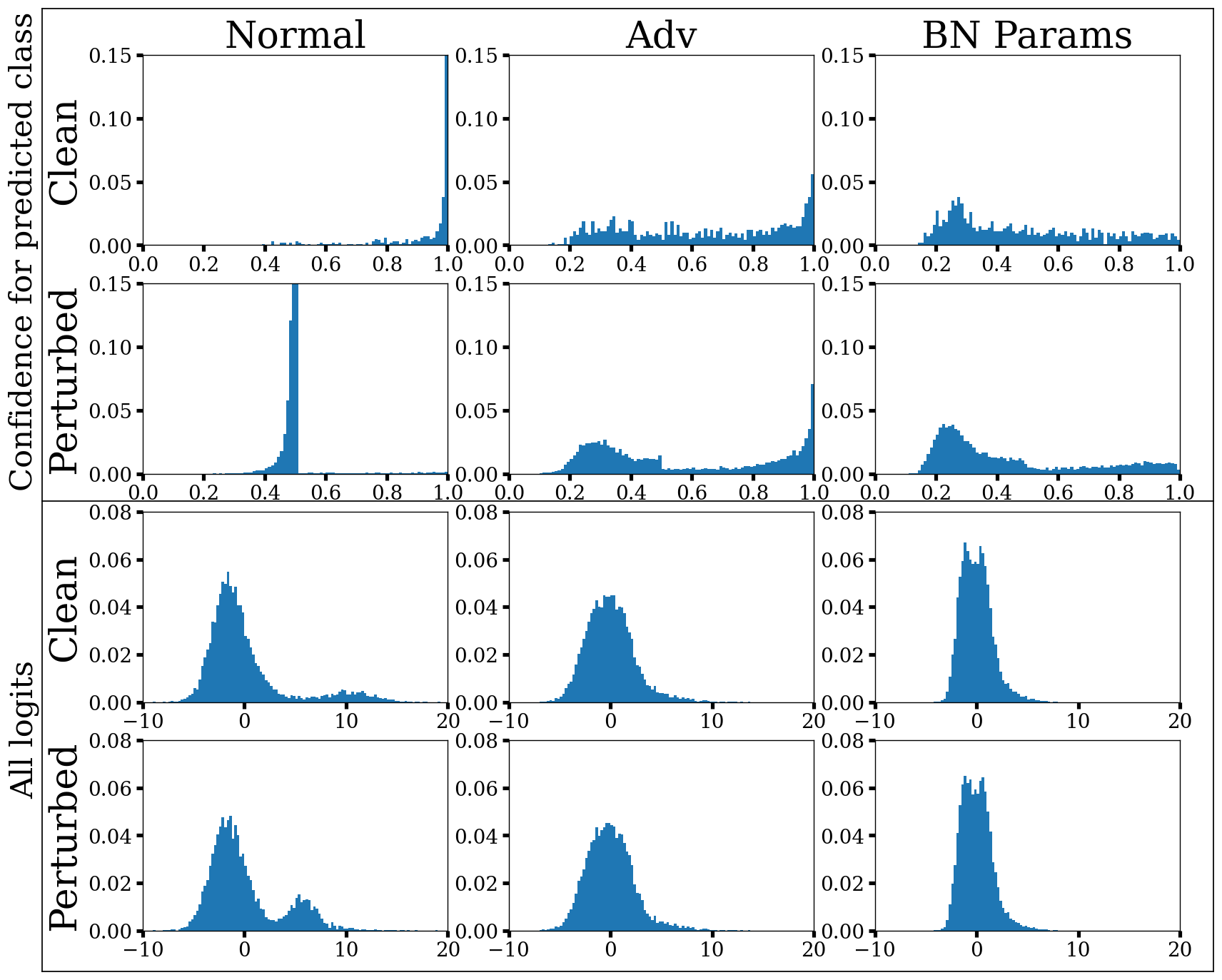}
    \vspace*{-9px}
    \caption{
    We plot confidence and logit histograms for \Normal, \Adv and \BNParams showing that \BNParams is able to follow the behavior of \Adv to a large extent, both on clean and adversarial examples.}
    \label{fig:logits}
    \vspace*{-0.625cm}
\end{figure}

\textbf{Non-Trivial Robustness:}
When adversarially fine-tune BN statistics \emph{and} parameters, in contrast, we are able to obtain non-trivial robust test error of $70.5\%$.
Again, the statistics are automatically updated through forward passes, while the parameters are trained using stochastic gradient descent.
At the same time, the clean error increases significantly from $4.11\%$ to $30.78\%$.
This indicates that the combination of updated statistics and adversarially learned parameters makes the difference.
To understand how the BN statistics and parameters change, we examined histogram plots as depicted in Figure \ref{fig:parameters}.
The distribution of the BN parameters, for examples, becomes slightly more concentrated around zero across most layers.
However, considering the BN parameters only, without the statistics can be misleading, as the histogram of the variance collapses also. 
Therefore, when analyzing the BN transformation, we view
BN as \emph{one} affine tranformation of the form  $f(z) = mz+b$. A simple rearrangement yields:
\begin{align}
    \textbf{BN}(z_i) = \left(\frac{\gamma_i}{\sqrt{\sigma_i^2 + \epsilon}} \right) z_i  + \left(- \frac{\gamma_i \mu_i}{\sqrt{\sigma_i^2 + \epsilon}} + \beta_i \right)\label{eq:normalized}.
\end{align}
In the following we refer to $m = \nicefrac{\gamma_i}{\sqrt{\sigma_i^2 + \epsilon}}$ as \emph{normalized} weight
and $b = - \nicefrac{\gamma_i \mu_i}{\sqrt{\sigma_i^2 + \epsilon}} + \beta_i$ as \emph{normalized} bias.
The histograms of $m$ for \Normal, \Adv and \BNParams can be found in Figure \ref{fig:parameters}.
The plots indicate that, compared to \Adv, the normalized weights of \BNParams have a lower mean in most layers.
In order to make this result crisp, we explicitly computed the mean of the normalized weights per layer, \cf Figure \ref{fig:meanweights}.
The average difference between the mean of the normalized weights is $-0.177$.
This is a strong indicator that \BNParams has to ``turn off'' part of the normally trained and thus fragile features in order to improve robustness.
While the BN statistics alone are not capable of achieving this, the BN parameters can be learned to remove specific features (\ie, channels for convolutional layers).
Additionally training the logit and first convolutional layer does not improve adversarial robustness significantly.
Finally, Figure \ref{fig:logits} shows that the logit distributions on clean and adversarial examples also starts to resemble those observed for \Adv.

\textbf{Adversarially Training BN Only:}
Following the idea of \cite{frankle2021training}, we also conducted experiments with only training the BN layers from scratch, but adversarially (\Adv + \BNOnly).
This means that the convolutional and fully-connected layers are initialized with random features and the BN layers attempt to utilize these features to obtain adversarial robustness.
Table \ref{table:main_summary} shows that we are unable to obtain non-trivial adversarial robustness based on random features.
Specifically, \Adv + \BNOnly achieves a clean error of $43.44\%$ (significantly worse than \BNParams), which
leads to the conclusion that random features do not contribute to robustness.

\section{Discussion and Conclusion}

Overall, our experiments allow to draw an interesting conclusion: while random features are not useful to learn robust classifiers, fragile features allow to achieve non-trivial robustness.
We suspect this to be possible by ignoring some fragile features.
Even though the adversarial robustness does not match the adversarially trained baseline, this result is significant because it was largely believed that fragile features are inherently not robust.
However, the remaining robustness gap also shows that new (robust) features are necessary for proper adversarial robustness.
These can only be learned using adversarial training from scratch.

{\small
\bibliographystyle{ieee_fullname}
\bibliography{main,david}
}

\end{document}

%% file: commands.tex
\usepackage{multirow}
\usepackage{nicefrac}
\usepackage{pifont}
\newcommand{\cmark}{\ding{51}}%
\newcommand{\xmark}{\ding{55}}%

\makeatletter
\DeclareRobustCommand\onedot{\futurelet\@let@token\@onedot}
\def\@onedot{\ifx\@let@token.\else.\null\fi\xspace}
\def\eg{e.g\onedot} 
\def\ie{i.e\onedot} 
 
\def\cf{cf\onedot}

\makeatother

\newcommand{\Normal}{\textsc{Normal}\xspace}
\newcommand{\Adv}{\textsc{Adv}\xspace}
\newcommand{\BNStats}{\textsc{BN Stats}\xspace}
\newcommand{\BNParams}{\textsc{BN Params}\xspace}
\newcommand{\BNOnlyParams}{\textsc{BN Only Params}\xspace}
\newcommand{\BNOnly}{\textsc{BN Only}\xspace}